# Soil moisture estimation from Sentinel-1 interferometric observations over arid regions

Kleanthis Karamvasis[a] and Vassilia Karathanassi[a]

[a]Kleanthis Karamvasis and Vassilia Karathanassi Laboratory of Remote Sensing, National Technical University of Athens, 9 Iroon Polytechniou Str., Zografou, 15780 Athens, Greece.

## ABSTRACT

We present a methodology based on interferometric synthetic aperture radar (InSAR) time series analysis that can provide surface (top 5 cm) soil moisture (SSM) estimations. The InSAR time series analysis consists of five processing steps. A co-registered Single Look Complex (SLC) SAR stack as well as meteorological information are required as input of the proposed workflow. In the first step, ice/snow-free and zero-precipitation SAR images are identified using meteorological data. In the second step, construction and phase extraction of distributed scatterers (DSs) (over bare land) is performed. In the third step, for each DS the ordering of surface soil moisture (SSM) levels of SAR acquisitions based on interferometric coherence is calculated. In the fourth step, for each DS the coherence due to SSM variations is calculated. In the fifth step, SSM is estimated by a constrained inversion of an analytical interferometric model using coherence and phase closure information. The implementation of the proposed approach is provided as an open-source software toolbox (INSAR4SM) available at www.github.com/kleok/INSAR4SM.

A case study over an arid region in California/Arizona is presented. The proposed workflow was applied in Sentinel-1 (C-band) VV-polarized InSAR observations. The estimated SSM results were assessed with independent SSM observations from a station of the International Soil Moisture Network (ISMN) (RMSE: 0.027 $m^3/m^3$ R: 0.88) and ERA5-Land reanalysis model data (RMSE: 0.035 $m^3/m^3$ R: 0.71). The proposed methodology was able to provide accurate SSM estimations at high spatial resolution (~250 m). A discussion of the benefits and the limitations of the proposed methodology highlighted the potential of interferometric observables for SSM estimation over arid regions.

1. Introduction

Soil moisture is considered a crucial environmental variable related to water, energy, and carbon cycles (Berg and Sheffield, 2018). In arid and semi-arid environments, which cover around 40% of global terrestrial surface, soil moisture exhibits significant spatio-temporal variability (Srivastava et al., 2021) and is a critical input to weather, climate, drought and flood modelling and forecasting (Jordan et al., 2020). In order to study the complex patterns of soil moisture, methodologies and surface soil moisture (SSM) products that exploit satellite earth observation data have been developed (Balenzano et al., 2021; Gao et al., 2017). However, the challenge of an operational soil moisture product with high spatial resolution and high spatiotemporal coverage remains elusive (Peng et al., 2021). Current and upcoming satellite missions offer opportunities to bridge the gap between soil moisture products and user/application requirements. SAR missions which can provide all-weather, day and night, global, and data with extensive and continuous coverage, have a critical role for high-resolution satellite SSM applications (Peng et al., 2021). Recent studies (Bürgi and Lohman, 2021; Mira et al., 2022a; Ranjbar et al., 2021) highlight the great potential of interferometric synthetic aperture radar (InSAR) time series for high-resolution and accurate SSM mapping over arid environments.

Time series InSAR is a well-established technique to estimate ground surface deformation (Yunjun et al., 2019). In order to extract deformation signals, other factors that can decorrelate SAR signals such as residual topography, atmospheric delay, vegetation and soil moisture variations have to be compensated (Zwieback et al., 2017). In most of the ground deformation studies, in order to overcome limitations associated with decorrelation, a methodology based on phase-stable point scatterers technique and/or distributed scatterers is implemented (Even and Schulz, 2018; Ferretti et al., 2011; Hooper et al., 2012). In those studies, soil moisture contributions are treated as noise and mitigated by applying filtering approaches. However, due to the observed relationship between SSM and InSAR observations, methodological approaches able to estimate SSM from InSAR phase and coherence information have been developed (Bürgi and Lohman, 2021; De Zan et al., 2014; Jordan et al., 2020; Mira et al., 2022a; Ranjbar et al., 2021; Zwieback et al., 2015).

The InSAR observables that are mainly used for SSM estimation are interferometric phase, coherence magnitude and phase closure. De Zan et al., (2014) proposed an analytical solution for bare lands based on plane waves and the Born approximation that models soil moisture as vertical complex wavenumbers and successively interferometric observables from L-band airborne SAR data. Coherence and phase closure observables were inverted to estimate soil moisture. Zwieback et al., (2015) assessed the relationship of the three aforementioned observables with the soil moisture using regression tools, for airborne and UAV L-band SAR. A positive relationship with increased sensitivity between soil moisture and interferometric phase, compared to the two other observables, was reported. Moreover, changes in soil moisture were found to be associated with a loss of coherence. Zwieback et al., (2017) highlighted the importance of phase referencing in order to estimate soil moisture. Phase referencing was found challenging using only InSAR observations in cases where displacement and soil moisture variations are correlated. De Zan and Gomba, (2018) used satellite L-band data and estimated successfully soil moisture at sub-kilometer scale by inversing phase closures constrained by coherence information. In this study, the need for evaluating the potential of phase closure for soil moisture estimation in the context of an operational approach and in shorter wavelengths was highlighted.

Michaelides, (2020) introduced a partially correlated interferometric model (PCIM) that considers the statistical properties of surface roughness and volume scattering. PCIM model yields a closer match to the simulation results than De Zan`s model (De Zan et al., 2014). Bürgi and Lohman, (2021); Jordan et al., (2020); Scott et al., (2017) introduced approaches able to generate accurate SSM proxies after precipitation events from InSAR coherence over arid and hyper-arid environments. Mira et al., (2022) proposed a two-step SSM estimation method for bare soils that uses C-band (Sentinel-1) SAR data. In the first step, the atmospherically corrected interferometric phase and coherence were calculated by exploiting nearby Global Navigation Satellite System (GNNS) observations. In the second step, the De Zan`s model were inverted and SSM were estimated with high accuracy (RMSE = 0.04 $m^3/m^3$) over bare soils. To the best of our knowledge, a limited amount of existing studies (Bürgi and Lohman, 2021; Mira et al., 2022a; Scott et al., 2017) are based on C-band InSAR datasets. We believe that the continuously increasing volume of C-band, mainly due to Sentinel-1 constellation, offers a great opportunity for exploring the potential of C-band data for SSM estimation. In this study, we developed a SSM estimation methodology for arid environments based on VV polarized Sentinel-1 InSAR time series. The implementation of the proposed approach is provided as an open-source software toolbox (INSAR4SM) available at www.github.com/kleok/INSAR4SM under GPL v3 license. The main innovations of the proposed method are:

- InSAR coherence and phase closure observables are calculated by forming distributed scatterers (Ferretti et al., 2011) and not over squares assuming homogeneity of scatterers (Bürgi and Lohman, 2021; Jordan et al., 2020).
- Use of external meteorological and SAR backscattering information for resolving coherence ambiguity which is reported at (De Zan and Gomba, 2018), in order to calculate ordering of SAR acquisitions based on their SSM levels.
- Constrained inversion by using the identified dry SAR acquisition. Improved initialization of the inversion by exploiting the relationship between coherence and SSM (Bürgi and Lohman, 2021; Jordan et al., 2020).

This paper is organized as follows. We first describe the methodological steps of the proposed approach (section 2). The study area and the datasets are presented in section 3. We then present the results together with the accuracy assessment (section 4). A discussion of the benefits and limitations of the proposed method is presented (section 5). We conclude at section 6.

2. Methodology

The required inputs for our methodology are a) a co-registered SLC stack and b) a meteorological dataset (precipitation, snow and temperature) for the same area and time considered. Currently, SLC stacks that are prepared using ISCE TopStack functionality (Fattahi et al., 2017) are supported by the INSAR4SM tool. The main methodological steps (Figure 1) are described as follows:

Step 1: Identification of driest SAR acquisition. Over arid regions, we assume that the majority of the selected SAR acquisitions correspond to low SSM conditions (Shao et al., 2021). Using the input meteorological dataset, we select the SAR acquisitions that are ice/snow free and are not related to precipitation activity. The driest SAR acquisition is selected as the one that yields the highest mean coherence in respect to the other selected images.

Step 2: Coherence and phase closure calculation. First, a SSM estimation grid is constructed according to specified grid size (e.g. 250 m). For each SSM grid cell, a DS is constructed from SLC SAR pixels that have similar intensity temporal patterns. A DS is a group of SLC SAR pixels that has lower noise in respect with a single SLC SAR pixel (Ferretti et al., 2011). The selection of those SLC SAR pixels is performed by using the non-parametric statistical Kolmogorov-Smirnov test (Ferretti et al., 2011). Then, the matrix $Z^{pxl}$ is formed by using the $l$ selected SLC SAR pixels at $p$ SAR acquisitions. We assume that the statistically homogeneous region of each SSM grid cell that consists of $l$ selected SLC SAR pixels, follow a zero-mean complex circular Gaussian model with sample covariance matrix $C$ (Ansari et al., 2018, 2017; Ferretti et al., 2011) that can be expressed by the following formula:

$$C = \frac{ZZ^H}{\sqrt{\|Z\|^2(\|Z\|^2)^T}} \qquad (1)$$

Where, $.^H$ indicates the Hermitian conjugation, $\|Z\|$ gives the row-wise Euclidean norm of the matrix Z. Power and division operations are elementwise. According to (Ansari et al., 2018), an element of $C$ at SAR acquisitions $n, m$ can be decomposed/considered as:

$$C_{nm} = \gamma_{nm} I_{nm} \qquad (2)$$

$$I_{nm} = e^{j\varphi_{nm}} \qquad (3)$$

Where, $\gamma$ is the magnitude of the interferometric coherence, $I$ the complex interferometric value and $\varphi$ the interferometric phase over the region that is covered by the $l$ selected SLC SAR pixels. Imaginary number is denoted with $j$. Phase closure $\Xi$ for $n, m, k$ SAR acquisitions according to (De Zan and Gomba, 2018) can be expressed by the following formula:

$$\Xi_{nmk} = \arg(I_{nm} I_{mk} I_{kn}) \qquad (4)$$

Step 3: Identification of "dry" SAR acquisitions. For each SSM grid cell, the ordering of SAR acquisitions based on SSM levels is calculated by exploiting the magnitude of the interferometric coherence and SAR backscattering. The ordering algorithm was based on the hypothesis that similar SSM levels yield high coherence values and different SSM levels yield low coherence values (Bürgi and Lohman, 2021; De Zan and Gomba, 2018). The magnitude of the interferometric coherence between SAR acquisitions can be affected by several factors such as vegetation changes, sediment transport, soil moisture changes or other types of changes that alter the composition of scatterers within a resolution pixel (Zebker and Villasenor, 1992). It is important to state that the contribution of vegetation on coherence losses is assumed zero because of the low vegetation coverage of arid regions (Bürgi and Lohman, 2021). This allows us to exploit the coherence information for ordering SAR acquisitions based on their SSM levels.

The identified driest acquisition from Step 1 is selected as a starting point. Then, based on the coherence values of the remaining acquisitions in respect to the driest acquisition we order every other SAR acquisition. Besides the coherence-based ordering, we calculate a backscattering-based ordering by assuming the positive monotonic relationship between SSM and SAR backscattering levels (Hoskera et al., 2020). In the end, the best ordering is considered the one that yields the steepest negative slope of coherence with respect to the distance between the temporal positions of ordered SAR acquisitions (De Zan and Gomba, 2018). After the ordering, the 30% of the driest

SAR acquisitions are considered to have low SSM and characterized as "dry". The abovementioned statement is considered valid, over arid regions and using a considerable length of SAR time series (Shao et al., 2021).

Step 4: Calculation of coherence due to SSM variations. In this step, for each SSM grid cell we calculate the coherence due to SSM variations. Coherence is related to several decorrelation sources with the following product form (Bürgi and Lohman, 2021; Guarnieri and Tebaldini, 2008; Jordan et al., 2020):

$$\gamma = \gamma_0 \gamma_p \gamma_{sm} \qquad (5)$$

Where, $\gamma_O$ is the background coherence due to surface roughness, slope, etc; $\gamma_P$ is the permanent coherence loss due to permanent changes due to precipitation events, erosion, deposition; and $\gamma_{sm}$ is the coherence loss due to SSM changes. The abovementioned expression is considered valid given that we operate over arid environments and for SAR stack with small spatial baseline separation (Bürgi and Lohman, 2021). Over arid regions, the variations of low vegetation cover are considered to have negligible impact in coherence and small perpendicular baselines mitigate geometric decorrelation that cause coherence loss.

Coherence loss due to SSM over "dry" SAR acquisitions is assumed to be negligible ($\gamma_{sm} = 1$). Then, $\gamma_O$ and $\gamma_P$ are estimated by considering the exponential model introduced by (Morishita and Hanssen, 2015) only over the coherence values calculated by the "dry" SAR acquisitions $\gamma_{dry}$.

$$\gamma_{dry} = (\gamma_O - \gamma_P) e^{-t/\tau} + \gamma_P \qquad (6)$$

Where, $\tau$ is the decorrelation rate and $t$ the time span between SAR acquisitions. After fitting the model for each SSM grid cell we have a single $\gamma_O$ value, a single $\tau$ value and a temporal function of $\gamma_p$. Then, for each combination of SAR acquisitions we calculate $\gamma_{sm}$.

Step 5: SSM inversion. For each SSM grid cell, an inversion of the analytical interferometric model proposed by (De Zan et al., 2014) is performed. The selected model provides a direct relationship between the SSM change and the coherence and phase closure information. The modelled complex interferometric value between $n, m$ SAR acquisitions reported in De Zan et al., (2014) is:

$$I_{nm}^{model} = \frac{1}{2j(k_n - k_m^*)} \qquad (7)$$

Where, $k$ is the complex wavenumber in the vertical direction and $j$ is the imaginary number. The model assumes that soil moisture variations cause different propagation of the SAR signals that can be described by the complex wavenumbers in the vertical direction. The soil is modeled as a uniform lossy dielectric layer and $k$ is described by the formula (De Zan and Gomba, 2018):

$$k = \sqrt{\omega^2 \mu \varepsilon} \qquad (8)$$

Where, $\omega$ is the angular frequency; $\mu$ is the magnetic permeability; and $\varepsilon$ the dielectric permittivity.

Dielectric permittivity is related to soil moisture and is given by the Hallikainen's empirical model (Hallikainen et al., 1985). The following cost function $L$ for all $n, m, k$ SAR acquisitions is considered:

$$L = \sum_{n<m<k} |\Xi_{nmk} - \Xi_{nmk}^{model}|_{2\pi}^2 + \sum_{n<m<k} (|\gamma_{sm,nm} - \gamma_{nm}^{model}| + |\gamma_{sm,mk} - \gamma_{mk}^{model}| + |\gamma_{sm,nk} - \gamma_{nk}^{model}|)^2 \quad (9)$$

Where, $\Xi$ is the observed phase closure; $\Xi^{model}$ is the modelled phase closure based on De Zan`s model; $\gamma_{sm}$ the coherence due to soil moisture variations; and $\gamma^{model}$ the modelled coherence based on De Zan`s model.

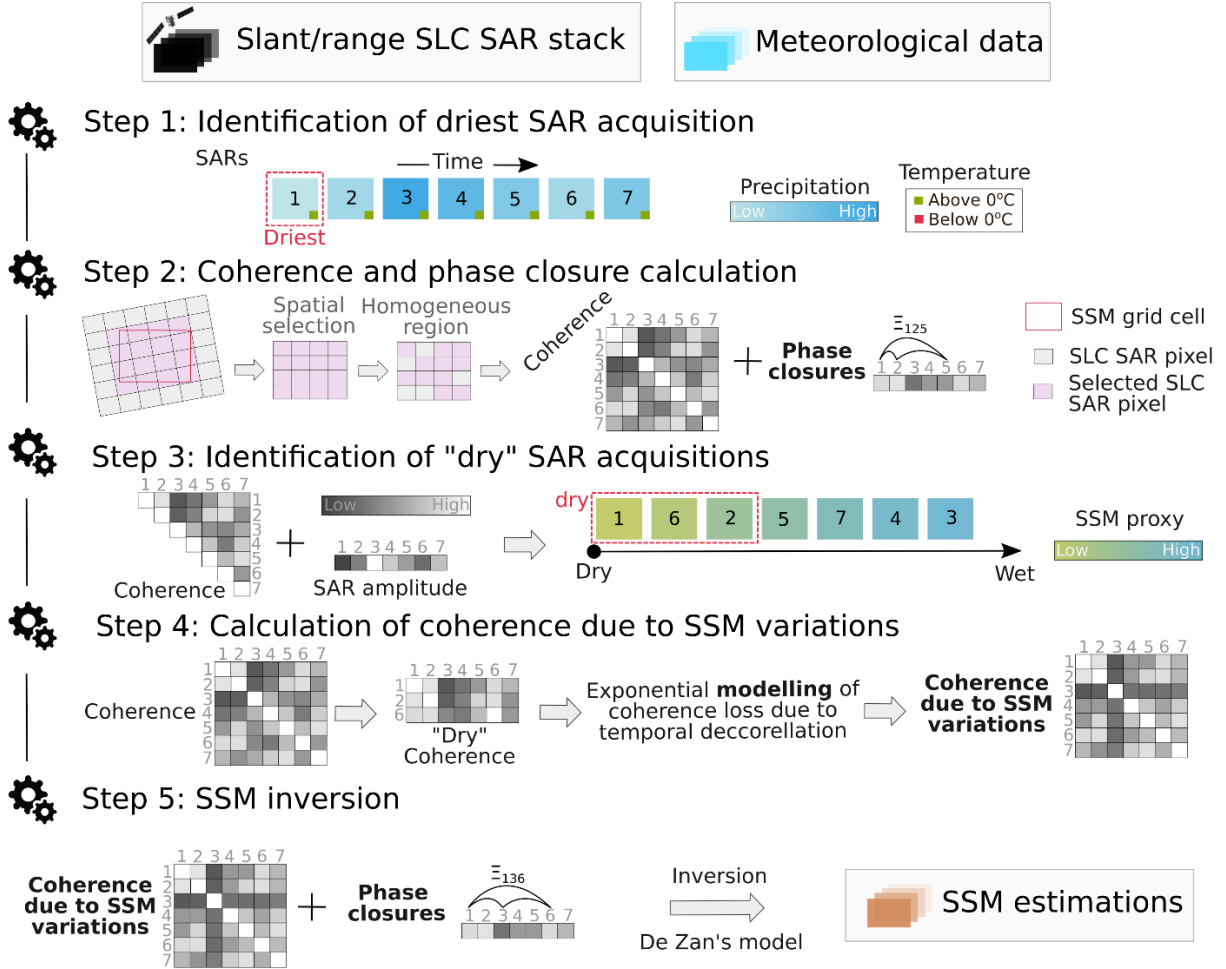

Figure 1 Methodological steps of the proposed approach

The Sequential Least SQuares Programming (SLSQP) solver ("Optimization and Root Finding," 2022) is selected due to its robust and fast performance. The solution space of the SSM values is from 0 to 0.5. In our implementation, the SSM value of the driest SAR acquisition identified at step 1 is set to 0.03 m³/m³, which we assume a reasonable value for arid environments. Next, the computed coherence due to soil moisture variations (output of step 4) over "dry" SAR acquisitions (output of step 3) was exploited to sort SAR acquisitions based on their soil moisture. Then, starting from the driest SAR acquisition, the initial SSM values of all other SAR acquisitions are calculated by the following formula which is based on Eq. 4 in (Bürgi and Lohman, 2021).

$$sm_i = \frac{sm_j}{\gamma_{sm,ij}}, \quad if \ sm_i < sm_j \quad (10)$$

Where, $i, j$ are indices for SAR acquisitions and $sm$ the SSM level of a SAR acquisition.

The inversion procedure was constrained in order to yield the SSM value of the driest SAR acquisition with an upper limit equal to the predefined SSM value of the driest SAR acquisition (0.03 m³/m³). The method may produce a soil moisture offset that is due to an error in the considered initial soil moisture value of the driest SAR acquisition.

3. Application

The proposed methodology is applied to an arid region located at California, USA (Figure 2). In particular, the area of interest is located at central Sonoran desert, in the Lower Colorado River Basin, and has low annual rainfall (50–300 mm) and high temperatures (Coppernoll-Houston and Potter, 2018). The soil type of the area of interest is Orita gravelly fine sandy loam with low runoff rates and high percolation rates (West et al., 2018). Regarding the land cover of the area of interest, the vegetation of the area of interest has low coverage and consists mainly of creosote bush (Larrea divaricata) and white bursage (Ambrosia dermosa) (Coppernoll-Houston and Potter, 2018). Moreover, anthropogenic surfaces such as highways, gravel roads, solar farms and agricultural fields can be found. The area of interest has broad nearly flat valley bottoms to high rocky mountain peaks.

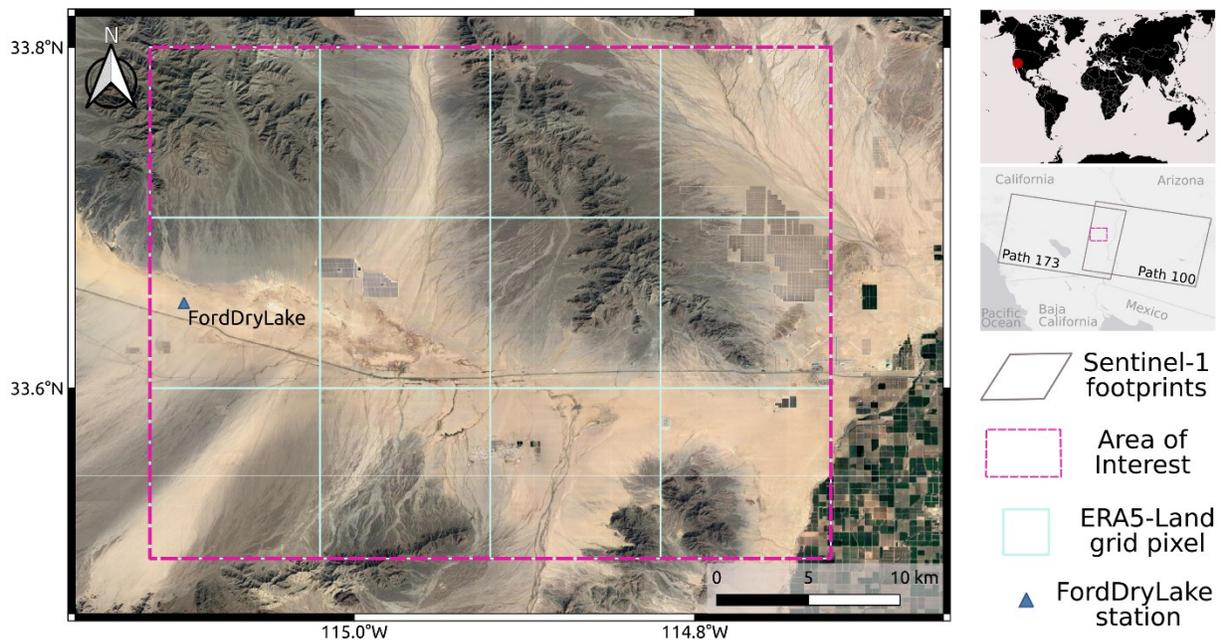

Figure 2 Area of interest and ground footprint of datasets used.

In the presented application, frequent (12-day repeats) Sentinel-1 (C-band) and ERA5-Land reanalysis meteorological data (Muñoz-Sabater et al., 2021) were used for estimating soil moisture in top 5 cm for bare soils over the area of interest. In our area of interest data from three orbits (173, 100, 166) of Sentinel-1 can be retrieved. For this study, we selected, 23 (orbit 173) and 22 (orbit 100) Sentinel-1 acquisitions acquired at about 14:30 UTC from July 2018 to March 2019. We dropped SAR acquisitions from orbit 166 because the acquisition time was during early morning (04.30 UTC). This decision is made because the thin water layer due to potential dew during the mornings affects the SAR signal and cannot be captured by soil moisture sensors used for validation (Mira et al., 2022b). Precipitation and temperature data from ERA5-Land model were used for identifying the driest SAR acquisition. For validation purposes in-situ soil moisture observations and soil moisture ERA5 model data were exploited. In-situ soil moisture measurements at depth of 5 cm from FordDryLake station of the SCAN network (Schaefer et al., 2007) that belongs to ISMN network (Dorigo et al., 2021, 2011) were used for assessing the performance of the proposed methodology. ERA5-Land reanalysis soil moisture data at depth of 0-7 cm were also used for quantitative assessment.

4. Results and accuracy assessment

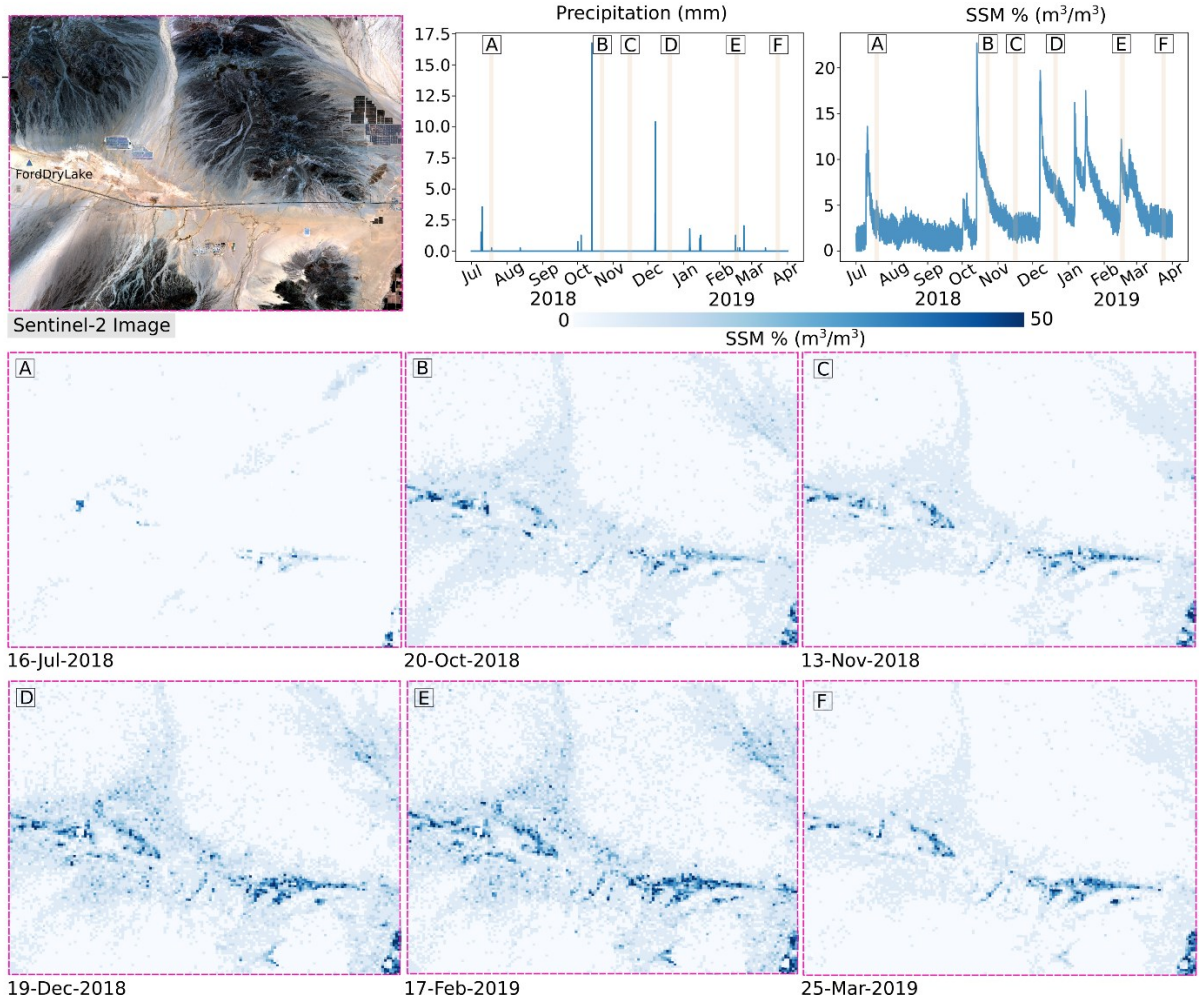

Figure 3 InSAR4SM SSM estimations

In Figure 3, the precipitation and soil moisture observations from the FordDryLake station are presented. Moreover, the SSM estimations at six selected time points on a 250 m SSM grid using the proposed methodology are presented. The 250 m SSM grid size was decided because it yielded the most accurate results after experimentation at the FordDryLake station. Based on the results, we can identify that valley regions experience more SSM variation in respect with the mountainous regions. Moreover, some regions tend to have higher SSM values in respect to other regions. For time points A, B and D in Figure 3, we observe that after rain events the SSM is increasing. The decay of SSM between time points (B to C) that we don't have rain events can also be observed. Based on the SSM estimations at time points E and F, we can also observe the temporal evolution of the SSM when no major rain events took place.

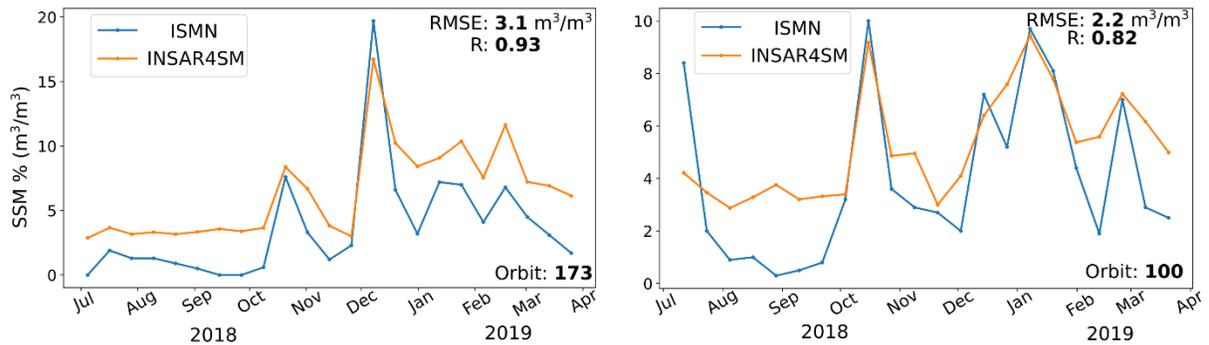

Figure 4 InSAR4SM SSM (volumetric %) estimations vs FordDryLake ISMN SSM (volumetric %) observations

In Figure 4, the soil moisture observations at 5 cm depth from FordDryLake ISMN station are compared with SSM estimations for orbits 173 and 100. For orbit 173, we have 23 SSM estimations with root mean square error (RMSE) of 0.031 m$^3$/m$^3$ and Pearson's correlation (R) of 0.93. For orbit 100, we have 22 SSM estimations with RMSE of 0.022 m$^3$/m$^3$ and R of 0.82. We can identify that in most cases, we have a positive monotonic relationship between the ISMN observations and InSAR4SM SSM estimations. In both orbits, the InSAR4SM was able to successfully capture the SSM increases due to rain events. However, in cases that we have a SSM decrease, SSM values are overestimated. We believe that this can be caused by the post-rain coherence loss that is not modeled by the proposed exponential temporal decorrelation model. It is also important to state, that the predefined SSM value for the driest acquisition plays an important role and can cause bias/offset effects visible in Figure 4.

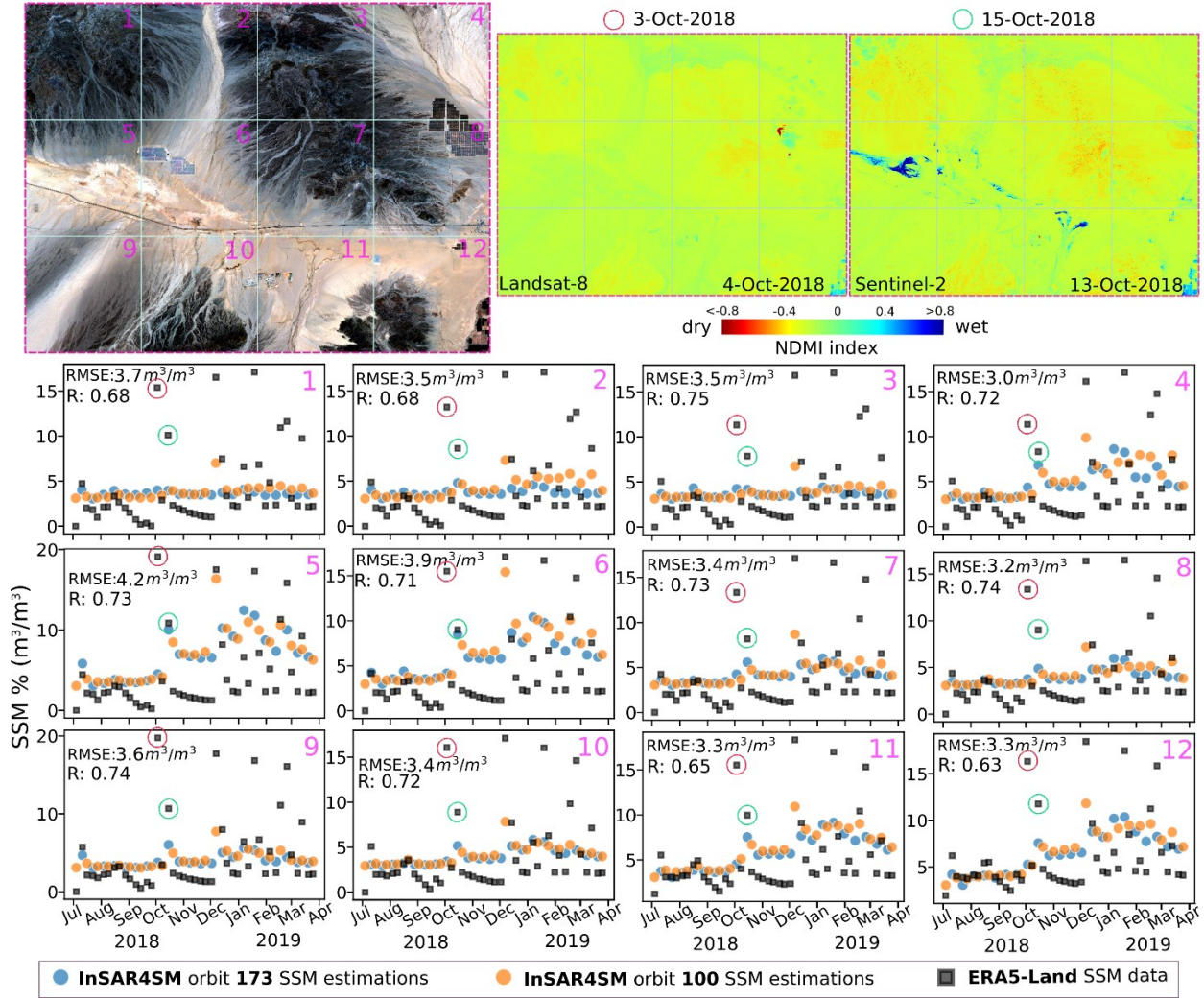

Figure 5 InSAR4SM SSM (volumetric %) estimations vs ERA5-Land soil moisture (volumetric %) estimations

In Figure 5, the comparison between the InSAR4SM SSM estimations and the ERA5-Land soil moisture data from 0 to 7 cm depth is presented. A consistency between the InSAR4SM SSM estimations from two different orbits can also be observed. For each ERA5-Land grid cell, we can observe the temporal pattern of the InSAR4SM SSM estimations in respect to ERA5-Land data. Different performance of InSAR4SM is observed for each ERA5-Land grid cell due to different land cover composition. In particular, InSAR4SM was not able to capture the SSM temporal pattern for ERA5-Land grid cells 1, 2, 3, 7, 8, 9 and 10 (figure 5) due to their rocky land coverage. Over the other ERA5-Land grid cells (4, 5, 6, 11 and 12), InSAR4SM was able to capture the SSM increases due to rain events. In all cases, InSAR4SM overestimated the SSM decreases after rain events in respect to ERA5-Land data.

For our accuracy assessment, it is important to also consider the accuracy of the ERA-5-Land soil moisture data (Muñoz-Sabater et al., 2021). In particular, based on the FordDryLake ISMN soil moisture observations (Figure 3) and Normalized Difference Moisture Index (NDMI) information (Costa-Saura et al., 2021) from multispectral data (Landsat-8 and Sentinel-2) (Figure 5) we identified an anomaly of ERA5-Land soil moisture data point (circled in red in Figure 5). In addition, we believe that the observed dispersion of ERA5-Land soil moisture data in 2019 (Figure 5) needs further investigation. We consider ERA5-Land soil moisture data the second best option after soil moisture observations from in-situ stations. After dropping the identified outlier, the mean RMSE value over the whole region of InSAR4SM estimations from orbits 173 and 100 is 0.035 $m^3/m^3$ with R value 0.71.

5. Discussion

InSAR4SM was able to produce SSM estimations with remarkable accuracy (~0.035 $m^3/m^3$) at high spatial resolution (250 m). Complex spatiotemporal patterns of SSM can be identified (Figures 3, 4) in our case study. Moreover, different orbits yielded consistent results without abnormal discontinuities (Figure 5). We believe that the good performance of InSAR4SM is due to three main methodological innovations. First, interferometric observables (InSAR coherence and phase closure) are calculated over homogeneous regions in terms of SAR backscattering (Ferretti et al., 2011). Due to the linkage between SAR intensity and SSM (Balenzano et al., 2021), by ensuring that the selected SAR pixels have similar intensity behavior we mitigate the risk to include scatterers that have different SSM patterns. This is considered a methodological improvement in respect to other approaches (Bürgi and Lohman, 2021; Jordan et al., 2020) that assume that all the SAR pixels inside a window share similar SSM patterns. Second, InSAR4SM exploits external meteorological data and SAR backscattering in order to identify the driest image, resolve coherence ambiguity, which is reported at (De Zan and Gomba, 2018) and order SAR acquisitions based on their SSM levels. Ordering is used for the identification of SAR acquisitions that have low SSM conditions. Those "dry" acquisitions are exploited to model coherence loss due to temporal decorrelation and calculate only the coherence information due to SSM variations. Third, SSM inversion procedure was improved because: a) an initialization based on the coherence-based SSM index (Eq. 10) was introduced in order to avoid ambiguities due to randomized initializations (De Zan and Gomba, 2018); b) a constraint based on the identified driest SAR acquisition was introduced; c) the decomposition of coherence and use only the coherence component related to SSM variations. Next, the main assumptions/limitations of the InSAR4SM approach are discussed.

A complex relationship between interferometric observables, SSM, and land coverage exists (De Zan and Gomba, 2018; Jordan et al., 2020; Scott et al., 2017). In this study, in order to estimate the coherence information due to SSM variations from the raw coherence a) we considered that coherence losses due to vegetation and geometric decorrelation are negligible and b) we modelled coherence mainly due to surface roughness, and permanent changes from precipitation events. It is important to state that land cover and ground surface roughness are assumed to be constant over time. However, in many cases this is not valid due to anthropogenic (e.g. constructions, mining, agricultural activity) and natural factors (e.g. vegetation changes, post-storm erosion and deposition of sediments). The coherence variations due to the abovementioned factors are not captured by the InSAR4SM`s exponential modelling of temporal decorrelation and SSM estimation errors are introduced. In our case study, this is mostly evident after rain events. In order to overcome this limitation, we believe that coherence loss due to post-storm erosion or deposition of sediments should be modelled. We believe, in order to better discriminate the coherence contribution from SSM variations it is important to consider more data related to topography, soil composition, temporal and spatial information of precipitation and land cover changes.

Another limitation is related to the complex relationship between different soil types and interferometric observables (Jordan et al., 2020). For this study, the soil composition of the FordDryLake station was used for the whole region.

This choice is considered reasonable due to the ongoing changes from fluvial sediment transport (Liu and Coulthard, 2015) and aeolian sand supplies (Arenas-Díaz et al., 2022) over desert environments. However, a lower performance of InSAR4SM mainly to rocky desert soils (Figure 5) was identified. We believe that incorporating soil properties information (Hengl et al., 2017) and studying interactions between soils and rocks can provide insights regarding the relationship between soil compositions and SSM estimation from interferometric observables.

After several experiments, we concluded that using a SSM grid of 250 m for our case study was suitable to provide accurate SSM estimations. The abovementioned spatial resolution is considered an improvement in comparison with backscattering approaches where 1x1km grid size is commonly used (Balenzano et al., 2021). The selected spatial resolution corresponds to about 12x50 C-band Sentinel-1 SLC resolution pixels. In order to enhance the signal quality and compute the interferometric observables, grouping of the statistically homogeneous pixels is required (Even and Schulz, 2018). In our view, spatial resolution can be improved if we ensure the quality of interferometric measurements. This can be achieved by either using SAR datasets with better spatial resolution or by using SAR data (e.g. L-band) with better signal-to-noise ratio. Further investigation is required in order to understand the relationship between spatial resolution and achieved accuracy by using SSM products with high spatial resolution.

Another assumption of this work is the validity of De Zan's interferometric model. Even though the model was developed for L-band data, it has been used also for C-band data (Mira et al., 2022a; Palmisano et al., 2022). The proposed InSAR4SM workflow can support L-band data from upcoming SAR missions (NISAR, ROSE-L). It is important to consider that L-band data can provide soil moisture estimations at deeper levels and have lower temporal decorrelation due to changes in the surface conditions (e.g. vegetation) over time. Even though, L-band interferometric observables are considered to have better quality in respect to C-band observables (Chen et al., 2021), over sandy regions L-band presents lower interferometric coherence than C-band (Wei and Sandwell, 2010). Finally, it would be interesting to evaluate the performance of a more sophisticated interferometric model (Michaelides, 2020).

Future work should include a) use of terrain information, land cover, rain rate, soil composition to understand and potentially model coherence loss due to post-storm erosion or deposition of sediments; b) use of soil properties from soil databases (Hengl et al., 2017) in order to better model the scattering behavior of soil; c) examine potential improvement of spatial resolution; d) expand our methodology in order to support upcoming L-band missions.

6. Conclusions

This paper provides a methodological approach and provides a description of the open-source software package InSAR4SM for soil moisture estimation over arid regions using InSAR and meteorological observations. Three methodological improvements were introduced and accurate SSM estimations at high spatial resolution over an arid region using Sentinel-1 data had been made. InSAR4SM contributes to better understanding the potential of interferometric observables for SSM estimation for arid regions. Currently, InSAR4SM can boost research initiatives and complement other soil moisture products. In the near future, it can support interferometric data from upcoming L-band SAR missions.